\documentclass[a4paper,twoside]{article}

\usepackage{epsfig}
\usepackage{subcaption}
\usepackage{calc}
\usepackage{amssymb}
\usepackage{amstext}
\usepackage{amsmath}
\usepackage{amsthm}
\usepackage{multicol}
\usepackage{pslatex}
\usepackage{apalike}
\usepackage[bottom]{footmisc}

\usepackage[breaklinks=true,bookmarks=false,backref=page]{hyperref}
\usepackage{amsfonts}
\usepackage{cite}
\usepackage{algorithmic}
\usepackage{epsfig}
\usepackage{graphicx}
\usepackage{textcomp}
\usepackage{caption}
\usepackage{array,multirow}
\usepackage{threeparttable}
\usepackage{booktabs}
\usepackage{bbm}

\usepackage{SCITEPRESS}     

\newcommand*\rot{\rotatebox{90}}

\begin{document}

\title{Hallucinated Heartbeats: Anomaly-Aware Remote Pulse Estimation}

\author{\authorname{Jeremy Speth, Nathan Vance, Benjamin Sporrer, Lu Niu, Patrick Flynn and Adam Czajka}
\affiliation{Computer Vision Research Laboratory, University of Notre Dame, Notre Dame, USA}
\email{\{jspeth,nvance1,bsporrer,lniu,flynn,aczajka\}@nd.edu}
}


\abstract{Camera-based physiological monitoring, especially remote photoplethysmography (rPPG), is a promising tool for health diagnostics, and state-of-the-art pulse estimators have shown impressive performance on benchmark datasets. We argue that evaluations of modern solutions may be incomplete, as we uncover failure cases for videos without a live person, or in the presence of severe noise. We demonstrate that spatiotemporal deep learning models trained only with live samples “hallucinate” a genuine-shaped pulse on anomalous and noisy videos, which may have negative consequences when rPPG models are used by medical personnel. To address this, we offer: (a) An anomaly detection model, built on top of the predicted waveforms. We compare models trained in open-set (unknown abnormal predictions) and closed-set (abnormal predictions known when training) settings; (b) An anomaly-aware training regime that penalizes the model for predicting periodic signals from anomalous videos. Extensive experimentation with eight research datasets (rPPG-specific: DDPM, CDDPM, PURE, UBFC, ARPM; deep fakes: DFDC; face presentation attack detection: HKBU-MARs; rPPG outlier: KITTI) show better accuracy of anomaly detection for deep learning models incorporating the proposed training (75.8\%), compared to models trained regularly (73.7\%) and to hand-crafted rPPG methods (52-62\%).}

\onecolumn \maketitle \normalsize \setcounter{footnote}{0} \vfill

\section{Introduction}

Remote vitals estimation is a growing field aiming to measure physiological signals from a camera. Perhaps the two most commonly estimated vitals are respiration rate and pulse rate, where algorithms predict a periodic waveform from a video. Several algorithms for estimating the blood volume pulse with remote photoplethysmography (rPPG) are robust to movement and even give error rates within FDA-approvable bounds on benchmark datasets~\cite{Chen2018,Speth_CVIU_2021}. 

While current estimators predict the correct signal if one exists, it is unclear if they are selective when there is no genuine pulse signal in the input. For instance: does the model generate a ``flatline" signal if no heartbeat exists in the video? Or does it generate a genuine-looking pulse waveform, hence giving improper feedback to the user (\eg medical practitioner) if the system is failing? These considerations generate {\bf four research questions} which we address in this paper:

\vspace{1.5ex}
\begin{itemize}
    \setlength\itemsep{2ex}
    
    \item [\textbf{(Q1)}] What do state-of-the-art rPPG models predict when a live subject is not in the video, or the pulse signal is too weak?
    
    \item [\textbf{(Q2)}] Can we build anomaly detection of abnormal pulse waveforms into existing deep learning rPPG estimators?

    \item [\textbf{(Q3)}] Can we train anomaly-aware rPPG models to reflect input signal quality in their predicted waveforms?

    \item [\textbf{(Q4)}] If the answer to Q3 is affirmative, how does anomaly-awareness affect performance of pulse rate estimation applied to genuine videos?

\end{itemize}
\vspace{1.5ex}

\begin{figure*}[t]
    \centering
    \includegraphics[width=\linewidth]{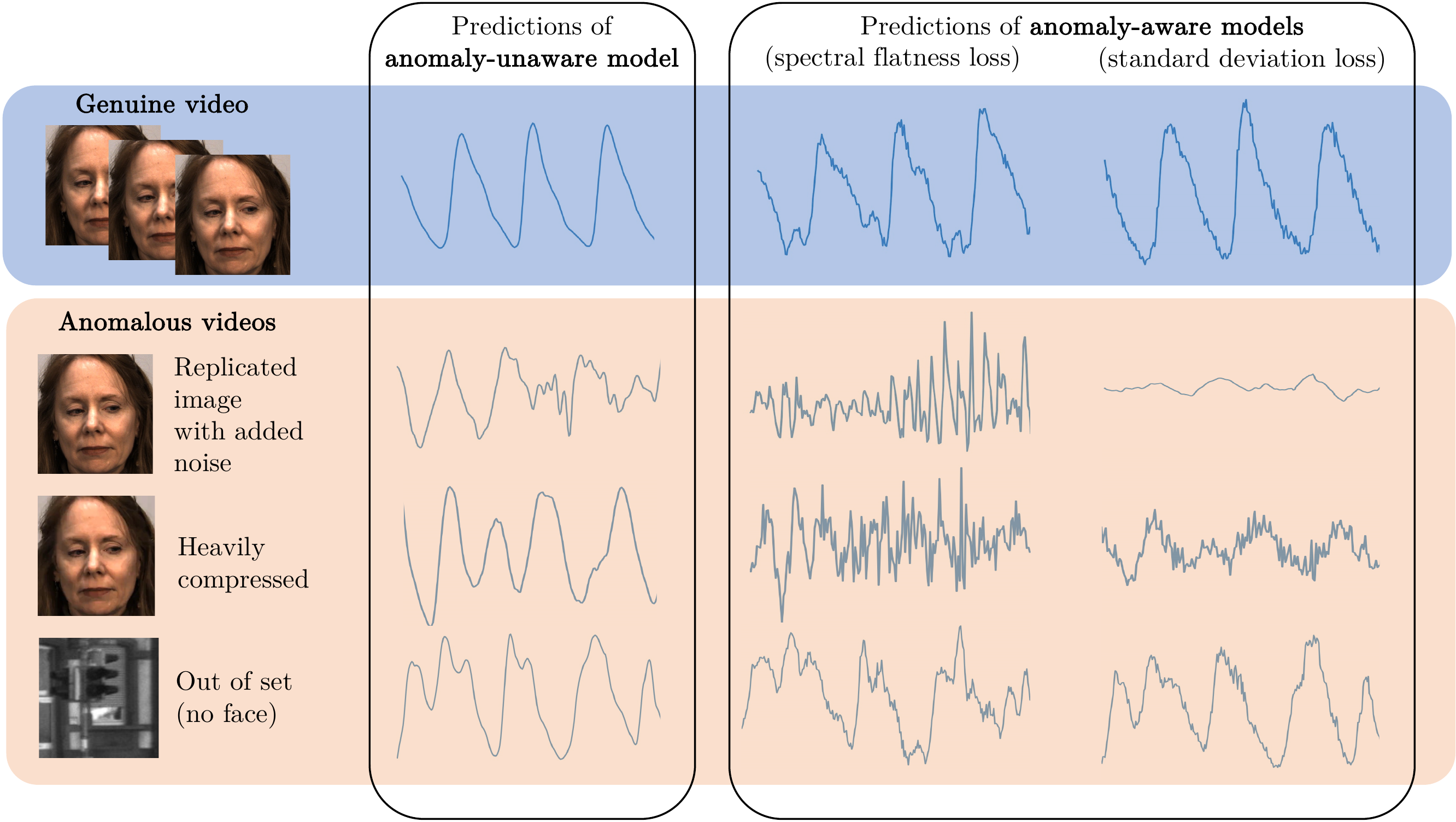}
    \caption{Given the black-box nature of modern, usually deep learning-based rPPG systems, can we trust that these models estimate correct waveforms for genuine inputs, and alert the examiner when abnormal samples are processed? Surprisingly, state-of-the-art solutions are able to ``hallucinate'' a realistically-looking pulse waveform from anomalous video that does not contain a human subject ({\bf bottom left} examples). To facilitate trustworthy vitals measurement, we propose {\it anomaly-aware} rPPG models that generate correct waveforms only for living individuals, and output low-quality, easy-to-detect signals for out-of-set samples. For real faces ({\bf top row} examples), the predicted signals are evaluated by comparing to ground truth collected from a fingertip pulse oximeter. Different training strategies, such as requesting flat spectrum ({\bf bottom middle} examples) or low standard deviation ({\bf bottom right} examples) of the predicted waveforms for out-of-set samples, along with closed- and open-set classifiers to easily detect ``hallucinated'' samples are also proposed.}
    \label{fig:teaser}
\end{figure*}

To answer the above questions we first show that state-of-the-art rPPG models (including deep learning approaches) are incapable of alerting the user of abnormalities in the predicted pulse waveform ({\bf re: Q1}). We then train binary classifiers to predict whether an input is anomalous from estimated waveform features on live videos and pulseless artificial videos. We also train one-class classifiers on waveform features from genuine videos to prepare the model for unseen data ({\bf re: Q2}). Next, we introduce various spectral regularization terms while training deep learning-based rPPG models that encourage a flat spectrum when the input videos do not contain a pulse signal ({\bf re: Q3}). Finally, we evaluate the proposed approaches on settings where the pulse is noisy or nonexistent, such as face presentation attacks, DeepFakes, compressed videos, rPPG attacks, and dynamic scenes (not containing faces at all) ({\bf re: Q4}).

We believe this work contributes to building trustworthy rPPG systems. The proposed models react better to unknown, noisy, and out-of-set signals by either alerting about abnormality or producing rPPG signals only for genuine inputs. To facilitate reproducibility and future research, we are releasing the source codes of the designed methods\footnote{\url{https://github.com/CVRL/Anomaly-Aware-rPPG}}.

\section{Related Work}

\subsection{Remote Photoplethysmography}
Remote Photoplethysmography (rPPG) is a technique for non-contact estimation of the blood volume pulse from reflected light. As the blood volume in microvasculature changes with each heart beat, the diffuse reflection changes due to the strong light absorption of hemoglobin. The observable changes are very subtle even in modern camera sensors, and we suspect they are  ``sub-pixel''~\cite{McDuff2021}. Initial approaches utilized only the green channel~\cite{Verkruysse2008}. Color transformation approaches~\cite{DeHaan2013,DeHaan2014,Wang2017,Wang2019} are still used as strong baselines due to their robustness and cross-dataset performance.

Some deep learning approaches regress the pulse rate directly without a waveform. Niu \etal~\cite{Niu2018,Niu2020} passed spatial-temporal maps to ResNet-18, followed by a gated recurrent unit to predict the pulse rate. While the model is accurate on benchmark datasets, it lacks any measure of confidence, and may produce a feasible pulse rate on anomalous inputs without feedback to the user.

The most common deep learning approaches regress the pulse waveform values over a video~\cite{Chen2018,Yu2019,Liu_MTTS_2020,Lee_ECCV_2020,Lu2021,Speth_CVIU_2021,Zhao_2021,Yu_2022_CVPR}. Several approaches use frame differences to estimate the waveform derivative~\cite{Chen2018,Liu_MTTS_2020,Zhao_2021}, which has the benefit of only requiring spatial models. Many other approaches leverage spatiotemporal features, and can process video clips end-to-end~\cite{Yu2019,Lee_ECCV_2020,Lu2021,Speth_CVIU_2021,Yu_2022_CVPR}. There are numerous advantages to producing a full waveform, including the ability to extract unique cardiac features such as atrial fibrillation~\cite{Liu_JBHI_2022,Wu2022}, and the ability to estimate noise as a proxy for model confidence.

\subsection{Deep Anomaly Detection}
This paper relates closely to anomaly detection by training with generated~\cite{Lee2018} or anomalous~\cite{hendrycks2019oe} inputs. Lee \etal~\cite{Lee2018} use a generative adversarial network to sample inputs near the data distributions boundary, then penalize high confidence on generated samples.
Hendrycks \etal~\cite{hendrycks2019oe} studied outlier exposure, where an entire out-of-distribution (OOD) dataset is introduced to the model during training. They use cross-entropy loss with the uniform distribution over classes as the target for OOD samples.
Our approach encourages a uniform distribution over the frequency domain of the estimated pulse waveform. We find it trivial to create anomalous samples that are spatially similar to the training data distribution, and we only use simple transformations to the original video dataset. This, to our knowledge, has never been explored in the rPPG context.

\section{Motivation}
As rPPG systems become more common in commercial products, it is important that they gracefully fail in unexpected situations, rather than giving incorrect vitals measurements. This paper presents a step towards incorporating the rPPG signal quality into the model's estimate. While we focus on detecting viable pulsatility for a global waveform, there are also several applications for quantifying pulsatility over the spatial dimension of a video. 

For example, Wang \etal~\cite{Wang2015,Wang2017LivingSkin} showed that local rPPG estimation can be used for segmenting skin pixels in video frames. Spatial measurement can also be used to measure blood perfusion in transplanted organs to verify that sufficient volumes of blood flow are flowing to the new tissue~\cite{Kossack_2022_CVPR}. Another impactful extension of current rPPG algorithms to the spatial domain is blood pressure estimation via pulse transit time~\cite{Wu2022,Iuchi_2022_CVPR}. Although current approaches manually segment regions of interest for estimating a pulse waveform, an end-to-end model that both segments the skin and estimates the local blood volume would be valuable. All of the aforementioned algorithms require models to only predict the pulse where it exists, so regions with tissue can be properly separated from the background.

\section{Problem Definition}
We first formulate the process for designing accurate pulse estimators. State-of-the-art methods regress the pulse waveform rather than the pulse rate. Given a video volume $X^{N\times W\times H\times C}$, the goal is to predict a real value at each frame corresponding to the true blood volume pulse, $Y^{N}$, where $N$ is the number of frames, $W$ and $H$ are frame width and height, and $C$ is the number of image channels. Recently, the task has been effectively modeled directly via spatiotemporal neural networks. Another common approach is to use frame differences to predict the first derivative of the waveform, which only requires processing two-dimensional input via common convolutional neural network architectures~\cite{Chen2018}.

Models estimate a waveform from cropped and rescaled face frame sequences. They are optimized via stochastic gradient descent to minimize the loss between the estimate and a ground truth pulse label. Common loss functions include mean square error (MSE) and negative Pearson~\cite{Yu2019} correlation. Training models with the above framework leads to highly accurate pulse rate estimators on videos of live subjects, and unpredictable ``pulse'' (yet realistically-looking, as depicted in Fig. \ref{fig:teaser}) signals estimated for videos not containing living subjects.

This paper thus explores the setting where the rPPG system should inform the end-user of a failure, by either generating unrealistic-looking waveforms or providing an additional signal-quality-related output for anomalous input videos.
We formulate this problem as our first research question {\bf Q1} listed in the introduction.
To answer this question, we pass video samples that are anomalous (\ie they do not contain a live human with a pulse) to our trained estimators and analyze the waveforms. Throughout the rest of the paper, we refer to samples containing a pulse as positive samples, and anomalous samples as negative. As demonstrated later in Sec. \ref{sec:resQ1}, {\bf we find that accurate pulse regressors such as color-based and spatiotemporal deep learning models are not necessarily effective liveness detectors}.

\section{Approach}
\begin{figure}
    \centering
    \includegraphics[width=\linewidth]{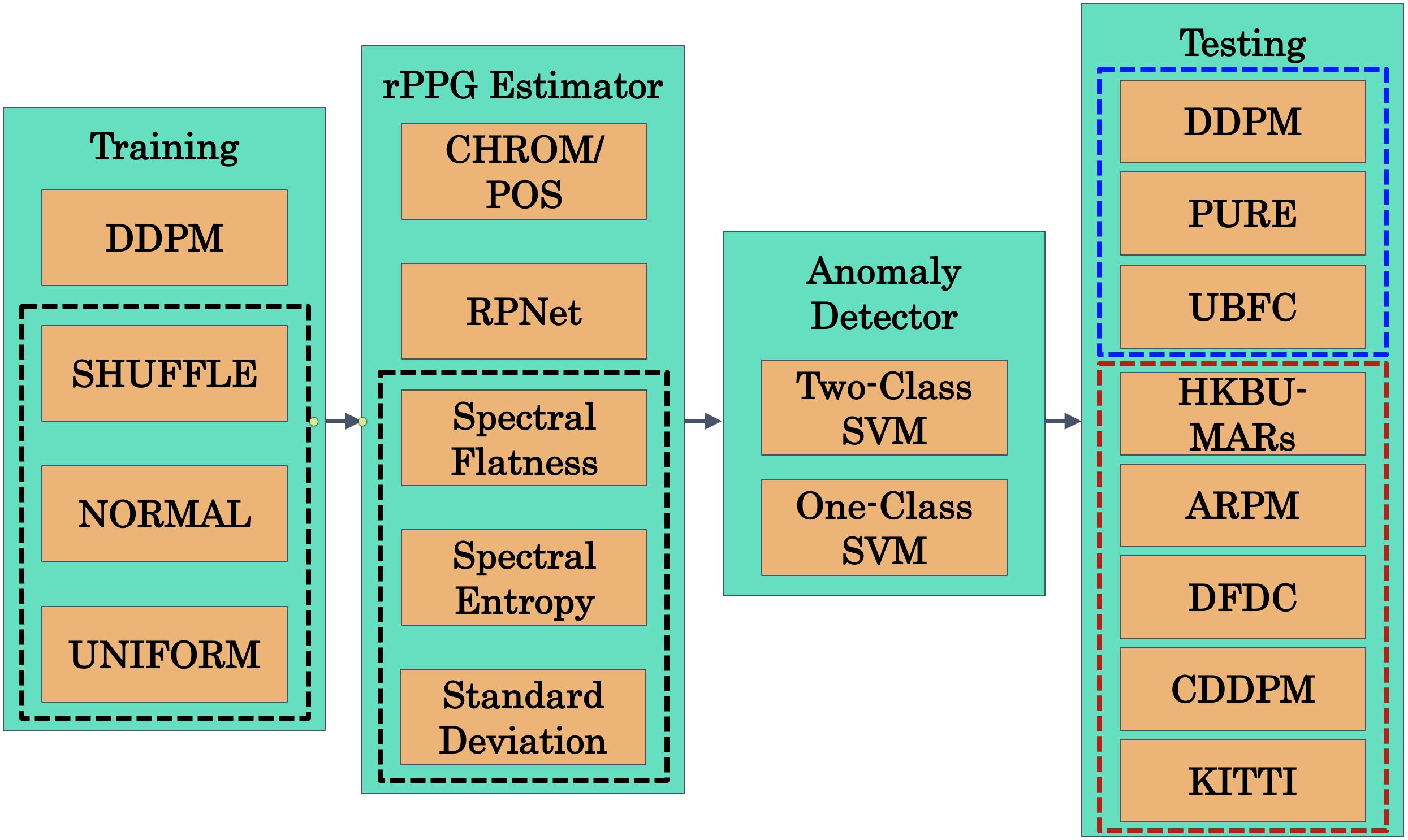}
    \caption{Models are trained on genuine DDPM and synthetic samples. Several features such as signal-to-noise ratio, standard deviation, and peak-to-peak distances are extracted from predicted waveforms to use as features for closed-set and open-set anomaly detection. We test on datasets with genuine samples (top 3) and anomalous samples (bottom 5).}
    \label{fig:exper_setup}
\end{figure}

To make the models ``aware'' of the input signal's quality, allow them to react differently for genuine and anomalous signals, and thus address our research question {\bf Q3}, we propose adding specially-crafted negative samples during training to reduce the sensitivity to pulseless input samples. The following sections describe how we designed samples without a pulse, and novel loss functions to penalize periodic predictions for such inputs. Also, to address our research question {\bf Q2}, we dicuss here binary classifiers for detecting anomalous inputs from the predicted waveform features. Figure \ref{fig:exper_setup}, along with the next subsections, summarize the training and testing experiment setups.

\subsection{Training dataset: Deception Detection and Physiological Monitoring (DDPM)}

DDPM dataset~\cite{Speth_IJCB_2021,Speth_CVIU_2021} consists of 86 subjects in an interview setting, where subjects attempted to respond to questions deceptively. Interviews were recorded at 90 frames-per-second for more than 10 minutes on average. Natural conversation and frequent head pose changes make it a difficult and less-constrained rPPG dataset.

\subsection{Designing Negative Samples}\label{sec:negative_samples_design}
In the true open-set regime, models are shown samples from unknown classes at inference time without having seen them during training. It is impossible to sample from the set of unknown unknowns, so training an open-set model in a supervised fashion is an incomplete modeling of the problem. However, it is straightforward to augment the existing video samples such that a true pulse signal does not exist. Assuming this heuristic approach to defining negative samples covers a sufficient portion of the negative space, we can artificially generate negative samples during training and train classifiers for binary liveness classification.

We define three different approaches for generating negative video samples taken from the DDPM dataset (illustrated at the left side of Fig. \ref{fig:exper_setup}):
\begin{enumerate}
    \item NORMAL: A single video frame is replicated over time with dynamic pixel-wise Gaussian noise added to the video.
    \item UNIFORM: As in (1), except uniform noise is added.
    \item SHUFFLE: The order of frames in a video is randomly shuffled.
\end{enumerate}
Hence, constructing useful negative samples for rPPG is mainly concerned with temporal dynamics of blood volume. In fact, the spatial contents of the input video can remain almost unaltered. This ensures the negative samples are spatially similar to positive samples, so we can sample near the boundary between the two classes. We use a standard deviation of 3 for sampling noise in NORMAL samples, and lower and upper bounds of -3 and 3, respectively, for sampling noise in UNIFORM samples. In all three approaches, a face is present in the input video, but the periodic signal is nonexistent. 

\subsection{Training with Negative Samples}
We trained the state-of-the-art rPPG model RPNet ~\cite{Speth_CVIU_2021} with the aforementioned negative samples to avoid periodic predictions in the absence of a pulse. Positive and negative samples were presented with equal probability to the model during training. Our complete loss formulation is given by a loss for positive samples, $\mathcal{L}_+$, and a loss for negative samples, $\mathcal{L}_-$:
\begin{equation}
\mathcal{L} = 
\begin{cases}
    \mathcal{L}_{+}(\hat{Y}, Y),& \text{if sample is positive with a pulse}\\
    \mathcal{L}_{-}(\hat{Y}),& \text{if sample is anomalous},
\end{cases}
\end{equation}
where $\hat{Y}$ is the model's waveform prediction and $Y$ is the ground truth pulse waveform. The positive loss minimizes the difference between the ground truth pulse of a positive sample and the prediction. The negative loss pushes predictions away from periodic features to make anomalous inputs easy to detect.
Note that the loss for negative samples only depends on the model's prediction, so the only requirement is finding samples known to be negative.

When considering signal quality, it is typically easier to formulate an objective in the frequency domain. We formulate two new FFT-based rPPG losses given a predicted waveform. Both losses use the normalized power spectral density:
\begin{equation}
    F(\hat{Y}) = \dfrac{\mathrm{FFT}(\hat{Y})}{\sum\limits_{i=1}^K \mathrm{FFT}(\hat{Y})},
\end{equation}
where $K$ is the number of frequency bins. To further constrain the frequencies for rPPG, we zero all frequencies lower than 40 bpm and higher than 240 bpm prior to normalization. We used PyTorch's FFT package to pass the gradient through the FFT operation. Full details for both training and inference of all pulse estimators can be found in the supplementary materials.

\subsubsection{Standard Deviation Loss}
Next, we explore penalizing the non-constant component of the predicted signal by using the standard deviation for negative samples. The goal of this loss is to reflect the input signal quality in the amplitude of the predictions and encourage the model to predict flatline signals. Positive samples still use the negative Pearson loss.

\subsubsection{Spectral Entropy Loss}
While the previous two losses penalize predictions in the time domain, we also use penalties in the frequency domain. A common method for measuring signal strength is signal-to-noise ratio (SNR), which compares power within a narrow band to the power outside that range. This approach is possible, because the blood volume pulse can typically be approximated by few frequencies. On the other extreme, a sample without a pulse should effectively return a white noise signal with a uniform spectrum. Shannon's entropy is a valuable metric for measuring the diversity of a distribution, which we believe is a good proxy for signal quality:
\begin{equation}
    \mathcal{L}_{-}(\hat{Y}) = - \dfrac{\sum\limits_{i=1}^K F(\hat{Y})\, \mathrm{log}(F(\hat{Y}))}{\mathrm{log}(K)},
\end{equation}
where $K$ is the number of frequency bins of the predicted waveform, $\hat{Y}$.
The loss penalizes waveforms with power concentrated amongst few frequencies, and encourages a white noise signal when no pulse exists in the input video.

\subsubsection{Spectral Flatness Loss}
Similarly to the our spectral entropy loss, we use spectral flatness~\cite{Dubnov2004} to penalize narrowband predictions on negative samples:
\begin{equation}
    \mathcal{L}_{-}(\hat{Y}) = \dfrac{1}{\sum\limits_{i=1}^K F(\hat{Y})} \, \mathrm{exp}\left(\frac{1}{K}\sum\limits_{i=1}^K \mathrm{log}(F(\hat{Y})\right),
\end{equation}
where $K$ is the number of discrete frequency bins, and $\hat{Y}$ is the estimated waveform for the anomalous sample.
Both the spectral entropy and spectral flatness losses range between 0 and 1.

\subsection{Anomaly Detection from Pulse Features}\label{sec:svm_features}
To address {\bf Q2}, we use handcrafted and interpretable features from estimated pulse waveforms for the prediction whether a sample is anomalous.
All features were calculated over a 10-second time window.
From the frequency domain, we extract the signal-to-noise-ratio (SNR), due its common use in rPPG experiments~\cite{DeHaan2013,Nowara_BOE_2021} and applications~\cite{Kossack_2022_CVPR} as a general signal quality metric.
We extract features related to the amplitude of the signal including the standard deviation ($\sigma$) and the envelope calculated from the Hilbert transform.

Additionally, we calculated several features from the peaks of negated signal (troughs).
Peaks were calculated with a Python implementation~\cite{ColakPyAMPD2016} of the automatic multiscale-based peak detection algorithm (AMPD)~\cite{ScholkmannAMPD2012}. 
First, we calculated the mean and standard deviation of the difference between consecutive peaks.
Next, we calculated the mean and standard deviation of the difference of differences, effectively examining the change in heart rate.
Finally, we calculated the root mean square of successive differences (RMSSD)~\cite{Shaffer2017}.

In general, SNR helps us understand how concentrated the signal power is in the frequency domain, and the amplitude features help measure how far the signal is from a ``flatline."
The peak-related features analyze the heart rate variability over short time frames. We concatenated all 8 features into a single feature vector for each 10-second window.

\begin{figure*}
    \centering
    \includegraphics[width=\linewidth]{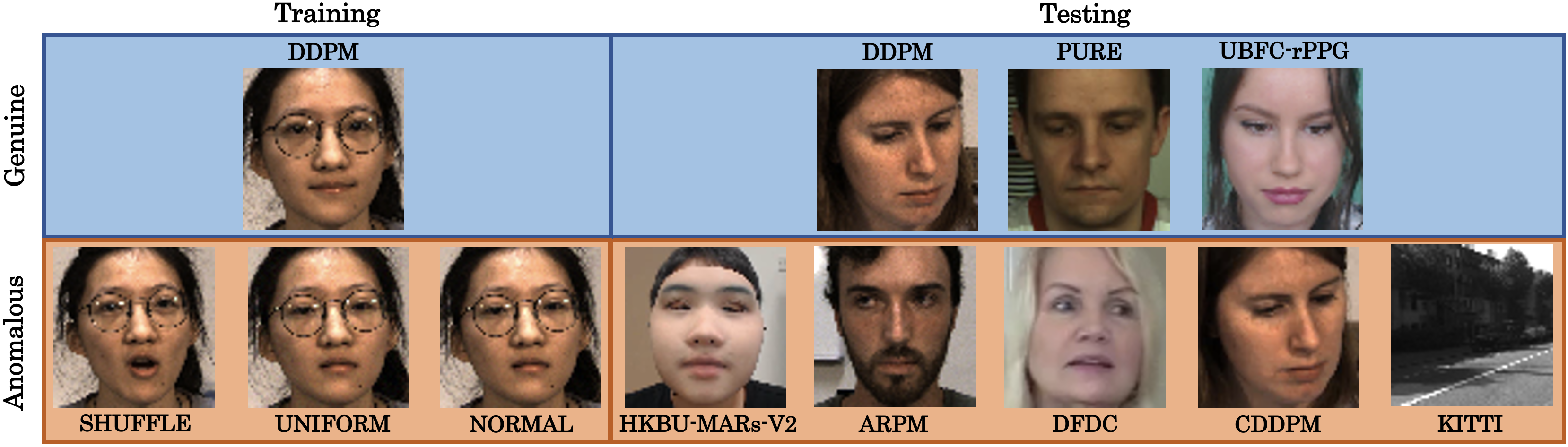}
    \caption{Frames of genuine samples from the DDPM training set ({\bf top left}) and benchmark rPPG datasets for testing ({\bf top right}). Synthetic frames transformed from the DDPM dataset ({\bf bottom left}) and testing frames from several anomalous datasets ({\bf bottom right}).}
    \label{fig:datasets}
\end{figure*}

We trained both one-class and two-class support vector machines (SVM) with radial basis function kernels on the aforementioned features for binary anomaly detection. The one-class SVMs were trained only on features from positive samples in the DDPM validation set. The two-class SVMs were trained on positive samples from the DDPM validation set and constructed negative samples described in section~\ref{sec:negative_samples_design}. Inclusion of a typical open-set classifier (one-class SVM) was to assess a general value of adding negative rPPG samples to the training regime. For both SVM architectures we used scikit-learn's default parameters. Figure \ref{fig:exper_setup} shows the overall pipeline for our approach and experiments.
\section{Experimental Evaluation}

\subsection{Pulse Test Datasets}

\noindent
\textbf{-- PURE}~\cite{Stricker2014}: PURE is a benchmark rPPG dataset consisting of 10 subjects recorded over 6 sessions. Each session lasted approximately 1 minute, and raw video was recorded at 30 fps. The 6 sessions for each subject consisted of: (1) steady, (2) talking, (3) slow head translation, (4) fast head translation, (5) small and (6) medium head rotations.

\noindent
\textbf{-- UBFC-rPPG}~\cite{Bobbia2019}: UBFC-rPPG contains 1-minute long videos from 42 subjects recorded at 30 fps. Subjects played a time-sensitive mathematical game to raise their heart rates, but head motion is limited during the recording.

\subsection{Pulseless Test Datasets}
We compiled several video datasets that do not contain a visible pulse to assess our approach. We mostly selected datasets that contain genuine or masked faces, such that the rPPG pipeline may not detect an anomalous input before passing video to the model. For a more extreme experiment, we used a dataset of dynamic scenes from a vehicle that do not contain faces or visible skin.

\noindent
\textbf{-- Compressed DDPM (CDDPM)}: Video compression is a well-studied challenge for rPPG~\cite{McDuff2017,Zhao_CVPRW_2018,Nowara_ICCVW_2019,Yu_ICCV_2019,Nowara_BOE_2021}. Most previous work has attempted to design models capable of robustly estimating the pulse on compressed videos. At high compression rates, however, estimation performance drops significantly, and usability of the system becomes questionable. To this end, we used the H264 video compression codec with a CRF value of 30. 

\noindent
\textbf{-- Adversarial Remote Physiological Monitoring (ARPM)}~\cite{Speth_2022_WACV}: Similarly to adversarial attacks in traditional biometric recognition, a recent work showed that rPPG systems are also prone to injection and presentation attacks that can change the predicted pulse rate. The dataset contains subjects sitting near an LED that projects a dynamic adversarial pattern on their skin. We consider the samples in this dataset to be negative, since a reliable rPPG system should detect an attack and warn the practitioner, rather than estimating the waveform.

\noindent
\textbf{-- DeepFake Detection Challenge (DFDC)}~\cite{Dolhansky_2019_dfdc, Dolhansky2020}: DeepFakes are realistic videos that change the identity of the original subject, while maintaining their actions. Significant efforts have been made to detect DeepFakes, as they pose a significant media threat. 

\noindent
\textbf{-- HKBU 3D Mask Attack with Real World Variations (MARs)}~\cite{Liu_2016_CVPRW}: We use version 2 of HKBU-MARs, which contains videos with both realistic 3D masks and unmasked subjects, to examine spatially realistic video with anomalous temporal dynamics. This is a valuable scenario to test, since the face masks are easily detected and landmarked by OpenFace~\cite{Baltrusaitis2018}, so the videos would be passed to the pulse estimator in a real system. Note that rPPG features were already shown to be useful on this dataset~\cite{Liu_WACV_2020}, but there evaluation is performed in a closed-set scenario, and does not evaluate models trained for robust pulse estimation.

\begin{figure*}[!ht]
    \centering
    \includegraphics[width=\linewidth]{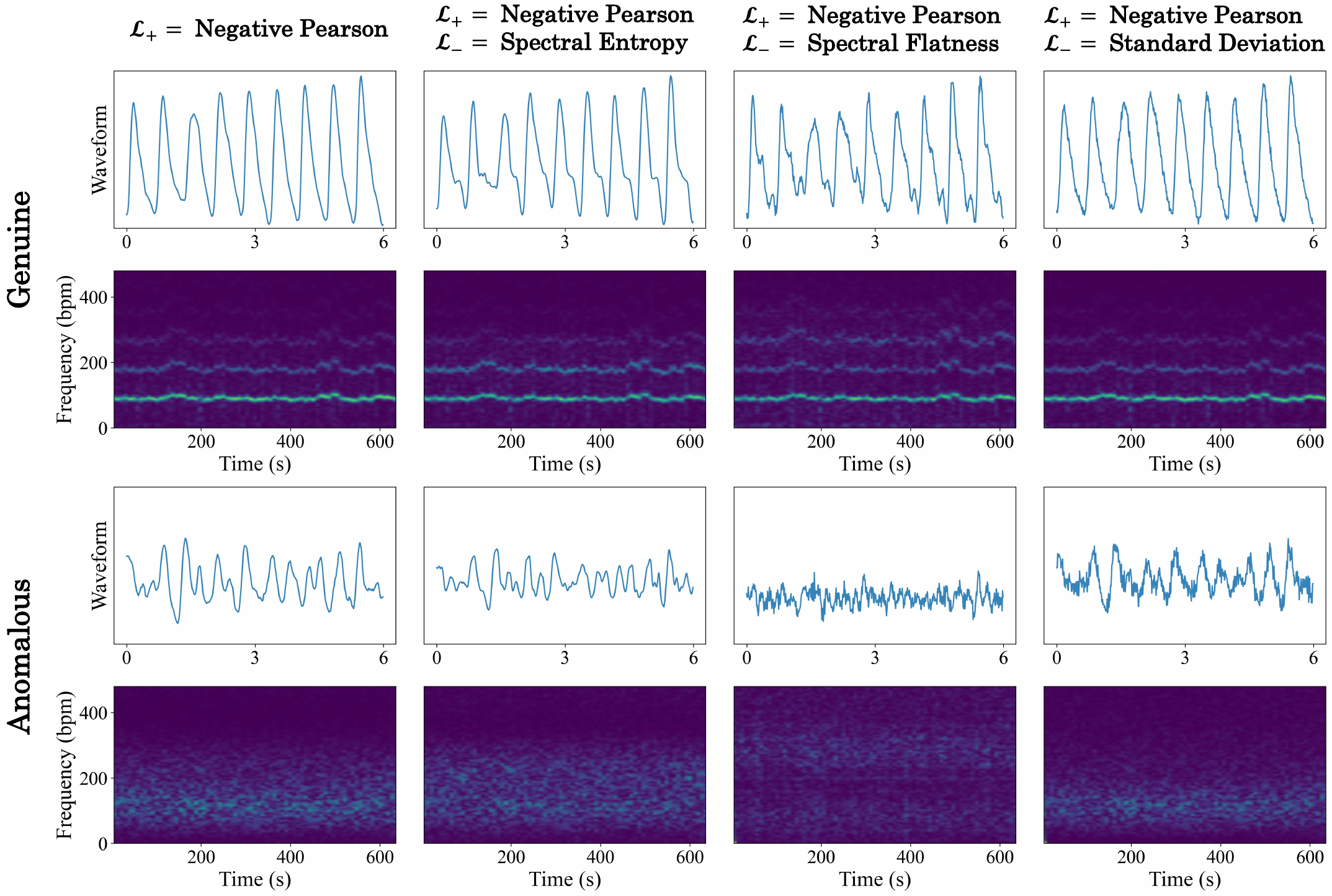}
    \caption{Waveform estimates over a 6-second window and spectrograms for an entire 600+ second video. The {\bf top} row shows predictions on a genuine video from the DDPM test set. The {\bf bottom} row shows predictions from a negative video generated by adding pixel-wise uniform noise to a single static frame from randomly selected from the same DDPM video. The first column represents an RPNet model \cite{Speth_CVIU_2021} trained only with positive samples and negative Pearson loss. The last three columns represent models trained with both positive and negative samples and the corresponding loss functions shown above.}
    \label{fig:spectrograms}
\end{figure*}

\noindent
\textbf{-- KITTI}~\cite{Geiger2013IJRR}: The KITTI dataset is a benchmark for autonomous vehicles, and contains recordings from several sensors attached to a vehicle. We used video from city and residential settings as an extreme case to test our approach, since they do not contain faces or skin pixels. We randomly selected a square region of interest with a minimum length of 64 pixels across all frames in each video.

\subsection{Evaluation Metrics}

\textbf{To answer Q1} we qualitatively examine waveforms from pulse estimators on anomalous data.

\textbf{To answer Q2 and Q3} we evaluate the SVM classifiers' binary predictions of whether an input video contains a pulse. We train both one-class and two-class SVMs on the set of 8 features described in section \ref{sec:svm_features}. We calculate the accuracy as the number of correctly classified frames over the total number of frames. Accuracy is calculated for each SVM on all datasets separately, and then combined to give an overall evaluation for the various domains.

\textbf{To answer Q4} we estimate pulse rates for the DDPM, UBFC-rPPG, and PURE physiological monitoring datasets. Pulse rates are computed as the highest spectral peak between 0.66 Hz and 4 Hz (40 bpm to 240 bpm) over a 10-second sliding window. The same procedure is applied to the ground truth waveforms for a reliable evaluation~\cite{Mironenko2020}. We apply common error metrics amongst rPPG research, such as mean error (ME), mean absolute error (MAE), root mean square error (RMSE), and Pearson correlation coefficient between frame-wise prediction and label pairs.
\section{Results}

\begin{table*}[!htb]
\setlength\tabcolsep{7pt}
\centering
\caption{Accuracy for anomaly detection from rPPG features on all datasets. {\bf RPNet} stands for off-the-shelf model and training, while three last columns correspond to the off-the-shelf RPNet architecture with the proposed training regimes.}
\renewcommand{\arraystretch}{1.1}
\resizebox{\textwidth}{!}{
\begin{tabular}{lc|rrrrrr}
\toprule
& \textbf{Dataset} & 
\begin{tabular}{@{}c@{}}\textbf{CHROM}\end{tabular} &
\begin{tabular}{@{}c@{}}\textbf{POS\phantom{i}}\end{tabular} &
\begin{tabular}{@{}c@{}}\textbf{RPNet\phantom{xx}}\end{tabular} &
\begin{tabular}{@{}c@{}}\textbf{RPNet +\phantom{xi}}\\\textbf{Entropy\phantom{xi}}\end{tabular} &
\begin{tabular}{@{}c@{}}\textbf{RPNet +\phantom{xxi}}\\\textbf{Flatness\phantom{xxi}}\end{tabular} &
\begin{tabular}{@{}c@{}}\textbf{RPNet\phantom{xxxi}}\\\textbf{+ $\sigma$\phantom{xxxi}}\end{tabular}\\
\midrule
\multirow{9}{*}{\rot{\underline{\bf One-Class SVM}}} & 
DDPM  & 41.96  & 43.60  & 51.96 $\pm$ 3.32  & 59.22 $\pm$ 1.21 & 41.65 $\pm$ 3.32\phantom{0} & 42.02 $\pm$ 2.53\phantom{0} \\
& PURE  & 4.60   & 0.00   & 0.12 $\pm$ 0.18   & 0.76 $\pm$ 1.02  & 3.16 $\pm$ 4.24\phantom{0} & 13.58 $\pm$ 5.73\phantom{0} \\
& UBFC  & 22.87  & 0.00   & 21.37 $\pm$ 9.59  & 11.07 $\pm$ 4.60 & 21.86 $\pm$ 21.17 & 57.07 $\pm$ 2.71\phantom{0} \\
& CDDPM & 37.34  & 59.24  & 88.44 $\pm$ 1.64  & 95.77 $\pm$ 0.77 & 71.21 $\pm$ 5.03\phantom{0} & 87.35 $\pm$ 4.06\phantom{0} \\
& ARPM  & 100.00 & 98.68  & 31.95 $\pm$ 1.77  & 40.64 $\pm$ 8.17 & 80.22 $\pm$ 6.44\phantom{0} & 78.31 $\pm$ 1.71\phantom{0} \\
& DFDC  & 47.67  & 50.01  & 49.81 $\pm$ 0.32  & 49.97 $\pm$ 0.37 & 49.10 $\pm$ 0.65\phantom{0} & 49.80 $\pm$ 0.30\phantom{0} \\
& MARs  & 52.98  & 50.00  & 64.37 $\pm$ 1.09  & 63.49 $\pm$ 1.43 & 54.92 $\pm$ 3.60\phantom{0} & 56.05 $\pm$ 3.35\phantom{0} \\
& KITTI & 100.00 & 100.00 & 100.00 $\pm$ 0.00 & 100.00 $\pm$ 0.00 & 98.18 $\pm$ 2.91\phantom{0} & 91.69 $\pm$ 2.78\phantom{0} \\
\cline{2-8}\\[-1.85ex]
& All & 45.04 & 48.88 & 56.77 $\pm$ 1.06 & 60.67 $\pm$ 0.47 & 52.65 $\pm$ 2.68\phantom{0} & \textbf{61.11} $\pm$ 0.55\phantom{0} \\
\bottomrule\\[-1.6ex]
\multirow{9}{*}{\rot{\underline{\bf Two-Class SVM}}} & 
DDPM  & 91.31 & 100.00 & 97.69 $\pm$ 0.96  & 98.92 $\pm$ 0.45    & 97.17 $\pm$ 2.84\phantom{0} & 94.26 $\pm$ 2.91\phantom{0} \\
& PURE  & 99.72 & 100.00 & 100.00 $\pm$ 0.00 & 100.00 $\pm$ 0.00 & 94.51 $\pm$ 5.55\phantom{0} & 75.64 $\pm$ 14.00 \\
& UBFC  & 99.44 & 100.00 & 99.20 $\pm$ 0.61  & 99.95 $\pm$ 0.06  & 96.90 $\pm$ 3.76\phantom{0} & 91.12 $\pm$ 7.33\phantom{0} \\
& CDDPM & 1.82  & 0.09   & 76.38 $\pm$ 5.10  & 80.60 $\pm$ 4.42  & 4.35 $\pm$ 7.30\phantom{0} & 47.12 $\pm$ 24.00 \\
& ARPM  & 2.73  & 4.31   & 13.01 $\pm$ 2.20  & 19.69 $\pm$ 2.50  & 2.42 $\pm$ 2.94\phantom{0} & 12.41 $\pm$ 6.30\phantom{0} \\
& DFDC  & 47.33 & 49.99  & 48.44 $\pm$ 0.48  & 48.03 $\pm$ 0.60  & 48.84 $\pm$ 1.00\phantom{0} & 49.30 $\pm$ 0.60\phantom{0} \\
& MARs  & 69.25 & 50.00  & 52.54 $\pm$ 1.27  & 49.94 $\pm$ 0.90  & 54.23 $\pm$ 4.33\phantom{0} & 68.63 $\pm$ 9.35\phantom{0} \\
& KITTI & 96.50 & 0.00   & 0.00 $\pm$ 0.00   & 0.00 $\pm$ 0.00   & 42.78 $\pm$ 37.03 & 45.48 $\pm$ 22.83 \\
\cline{2-8}\\[-1.85ex]
& All   & 62.03 & 52.11 & 73.70 $\pm$ 1.23 & \bf{75.78 $\pm$ 1.32} & 56.68 $\pm$ 2.71\phantom{0} & 65.87 $\pm$ 7.21\phantom{0} \\
\bottomrule
\end{tabular}
}
\label{tab:svm_results}
\end{table*}

\subsection{Addressing Research Question Q1 (Predicted Waveforms for Anomalous Samples)}\label{sec:resQ1}
The bottom row of Fig.~\ref{fig:teaser} shows predictions from a standard RPNet model and anomaly-aware RPNet models. The top waveforms are from a DDPM sample, and all show a clear pulse signal. The bottom waveforms are from a KITTI sample of city driving. The prediction from the standard RPNet model looks visually similar to that of a blood volume pulse. In fact, the model's priors are so strong that it even adds a dicrotic notch to the third cycle. The anomaly-aware RPNet model trained with the spectral flatness penalty predicts a wideband signal without strong frequencies in the typical pulse range. The anomaly-aware RPNet model trained with standard deviation penalty predicts a low amplitude signal compared to that of the genuine prediction. {\bf To answer Q1, spatiotemporal deep learning models can produce genuine-looking waveforms from anomalous inputs, even adding distinct features such as the dicrotic notch.}

Additionally, the bottom row of Fig.~\ref{fig:spectrograms} shows estimated waveforms for a 6-second segment, and periodograms for several minutes from the original and regularized RPNet models on a still face frame with additive uniform noise. As shown in the periodogram, the original model occasionally estimates periodic components between 50 and 300 bpm, and effectively bandpass filters the signal. When training with spectral flatness penalty for negative samples, the model uniformly spreads the signal strength over all frequencies, giving a white noise signal. The spectral entropy model allows slightly higher frequency components than the original model. The standard deviation model allows for high frequency components as well, but surprisingly keeps a somewhat narrower frequency band between 50 and 180 beats per minute. Visually, the waveforms for the regularized models look more unrealistic than the original model, and correctly propagate input errors to the prediction.

\subsection{Addressing Research Question Q2 (Anomaly Detection On Top of Existing rPPG Estimators)}
The first three columns in Table~\ref{tab:svm_results} show the one-class and two-class SVM anomaly detection results for baseline pulse estimators. Features from the original RPNet result in an overall accuracy of 73.70\% with two-class SVMs. CHROM gives the highest accuracy of the color-transformation approaches with 62.04\%. POS gives the worst baseline performance, which we attribute to the lack of bandpass filtering, allowing for high frequency signals to occur regardless of the video input.

One-class classification is difficult, since only features from live samples in DDPM were used to fit the classifier, and it is an open-set problem. In the three upper-leftmost columns, the color transformation methods achieves lower than 50\% accuracy on the combined data, but the RPNet model achieves 56.77\% accuracy. {\bf To answer Q2, the color transformation methods struggle to extract useful features for anomaly detection, and the deep learning model (RPNet) with unmodified training is still relatively poor at propagating input signal quality to the waveform prediction.}

\subsection{Addressing Research Question Q3 (Anomaly Detection From Anomaly-Aware rPPG Estimators)}
Using two-class SVMs, the anomaly-aware model trained with spectral entropy gives the highest accuracy for detecting anomalous video inputs. Comparing to the original training regime for RPNet, we find a 4\% and 2\% increase in accuracy for one- and two-class SVMs with our proposed training strategy.
The one-class classifiers perform worse overall than the two-class classifiers, but the proposed training strategy still gives improvements over the original RPNet model. Standard deviation loss provides the richest features for one-class classification. We see that detecting anomalous inputs from waveforms is still a difficult task. {\bf Overall, the answer to Q3 is affirmative: the proposed training regimen results in more discriminative features for anomaly detection.}

\subsection{Addressing Research Question Q4 (Performance Of Anomaly-Aware Models For Live Subjects)}
Table \ref{tab:pulse_results} shows pulse rate performance across the RPNet models and baseline color transformation methods. The deep learning estimators outperform the baselines on DDPM, since they were trained on data in the same setting. For PURE, the baselines give the lowest error rates, and the RPNet models transfer poorly from DDPM. The poor performance could be explained by the low average pulse rate in PURE compared to DDPM, which is reflected in the strong bias from all deep learning models. Performance is relatively stable across deep learning and baseline models on the UBFC-rPPG. The vanilla RPNet model trained with simple negative Pearson loss on DDPM transfers well, giving a mean absolute error of 1.46 bpm.

Across all datasets, the model penalized with standard deviation gives the most accurate pulse rates. The deep learning models do not provide a significant improvement over baseline color transformation methods in cross-dataset testing. However, models trained with our technique have low errors when examining both the within-dataset and cross-dataset evaluations. {\bf Therefore, the answer to Q4 is that anomaly-aware training with negative samples does not harm pulse rate estimation and may even improve performance.} We believe this is an encouraging finding, and additional model regularization techniques for rPPG should be explored further. 

\begin{table}[!htb]
\setlength\tabcolsep{2.7pt}
\centering
\caption{Pulse rate estimation performance for baseline and spatiotemporal methods on all pulse datasets.}
\resizebox{\columnwidth}{!}{
\begin{tabular}{lccccc}
\toprule
&& {\bf ME} & {\bf MAE} & {\bf RMSE} & {\bf $r$}
\\
\midrule
\multirow{6}{*}{\rot{\underline{DDPM}}}
& CHROM        & 8.68 & 13.04 & 28.49 & 0.56\\
& POS          & 4.21  & 8.88  & 23.67 & 0.69\\
& RPNet & -1.91 $\pm$ 0.13 & \bf{3.41 $\pm$ 0.11} & \bf{12.38 $\pm$ 0.27} & \bf{0.91 $\pm$ 0.00}\\
& Entropy & -1.26 $\pm$ 0.45 & 4.06 $\pm$ 0.38 & 13.62 $\pm$ 0.68 & 0.89 $\pm$ 0.01\\
& $\sigma$ & -1.39 $\pm$ 0.35 & 3.75 $\pm$ 0.24 & 13.07 $\pm$ 0.58 & 0.90 $\pm$ 0.01\\
& Flatness & \bf{0.73 $\pm$ 2.18} & 7.00 $\pm$ 2.60 & 19.45 $\pm$ 4.61 & 0.76 $\pm$ 0.11\\
\midrule
\multirow{6}{*}{\rot{\underline{PURE}}}
& CHROM        & {\bf -0.02}  & {\bf 0.73}  & {\bf 2.14}  & {\bf 1.00}\\
& POS          & 0.13   & 0.77  & 3.84  & 0.99\\
& RPNet & -9.64 $\pm$ 2.73 & 13.21 $\pm$ 2.74 & 25.72 $\pm$ 3.61 & 0.60 $\pm$ 0.06\\
& Entropy & -6.74 $\pm$ 1.76 & 11.01 $\pm$ 1.02 & 25.73 $\pm$ 1.07 & 0.54 $\pm$ 0.02\\
& $\sigma$ & -5.34 $\pm$ 1.08 & 10.74 $\pm$ 2.48 & 22.21 $\pm$ 3.20 & 0.60 $\pm$ 0.11\\
& Flatness & -3.90 $\pm$ 5.11 & 9.41 $\pm$ 3.43 & 23.33 $\pm$ 7.58 & 0.62 $\pm$ 0.15\\
\midrule
\multirow{6}{*}{\rot{\underline{UBFC-rPPG}}}
& CHROM & 2.12  & 2.64 & 10.37 & 0.85\\
& POS   & 1.44  & 2.06 & 8.85  & 0.89\\
& RPNet & 0.18 $\pm$ 0.28 & \bf{1.46 $\pm$ 0.52} & \bf{5.64 $\pm$ 1.59} & \bf{0.94 $\pm$ 0.03}\\
& Entropy & \bf{0.15 $\pm$ 0.41} & 1.76 $\pm$ 0.77 & 7.68 $\pm$ 2.23 & 0.90 $\pm$ 0.05\\
& $\sigma$ & 0.63 $\pm$ 0.46 & 2.71 $\pm$ 1.48 & 8.98 $\pm$ 3.24 & 0.87 $\pm$ 0.08\\
& Flatness & -1.01 $\pm$ 4.91 & 5.41 $\pm$ 3.31 & 16.97 $\pm$ 7.55 & 0.69 $\pm$ 0.17\\
\midrule
\midrule
\multirow{6}{*}{\rot{\underline{\bf All}}}
& CHROM        & 4.82       & 7.37       & 20.87       & 0.74\\
& POS          & 2.46       & 5.15       & 17.46       & 0.82\\
& RPNet & -3.79 $\pm$ 0.87 & 5.92 $\pm$ 0.75 & 16.79 $\pm$ 1.52 & 0.83 $\pm$ 0.03\\
& Entropy & -2.61 $\pm$ 0.52 & 5.67 $\pm$ 0.50 & 17.40 $\pm$ 0.75 & 0.82 $\pm$ 0.02\\
& $\sigma$ & -2.16 $\pm$ 0.47 & \bf{5.61} $\pm$ 0.44 & \bf{15.91 $\pm$ 0.82} & \bf{0.84 $\pm$ 0.02}\\
& Flatness & \bf{-0.98} $\pm$ 3.26 & 7.40 $\pm$ 2.03 & 20.82 $\pm$ 3.92 & 0.74 $\pm$ 0.08\\
\bottomrule
\end{tabular}
}
\label{tab:pulse_results}
\end{table}

\section{Discussion}

\subsection{Do all models ``hallucinate" pulse waveforms?}
We show that spatiotemporal deep learning models can produce genuine-looking waveforms that do not exist in the input. Note that this problem will only occur in models that operate over the time dimension. For one-dimensional temporal neural networks~\cite{Wu2022}, the model could similarly extract a periodic signal within physiological bounds. Two-dimensional neural networks that estimate the pulse derivative~\cite{Chen2018,Liu_MTTS_2020} avoid the problem, since the model treats each time step independently.

\subsection{Other Sources of Artificial Frequencies}
Spatiotemporal and temporal models typically consume short overlapping segments from a video, and the predictions are ``glued" together with techniques such as overlap-adding~\cite{DeHaan2013}. Our initial experiments exposed artificial frequencies that were added to our predictions due to the default padding parameters in PyTorch. In the default 3-D convolution operation, the input video clips were being padded with zeros in the time dimension, which created an artificial temporal response from our model even on constant input videos. We recommend edge-based padding to mitigate this problem.
\section{Conclusions}
We present the first experiments to explore how spatiotemporal networks for rPPG behave when given anomalous video inputs, and the first rPPG models trained to appropriately react to anomalous input signals. Spatiotemporal networks learn such strong priors on the shape of blood volume pulse waveforms that they can ``hallucinate" a genuine-looking waveform when no live subject exists in the input video. To mitigate this problem, we propose a new training regimen for spatiotemporal models. We find that penalizing the model for predicting periodic signals on inputs without a human pulse yields more trustworthy predictions. Our experiments showed that features extracted from the proposed models were more powerful for detecting anomalous videos and even gave lower error rates for pulse rate estimation applied to genuine videos.
\section*{\uppercase{Acknowledgements}}
This research was sponsored by the Securiport Global Innovation Cell, a division of Securiport LLC. Commercial equipment is identified in this work in order to adequately specify or describe the subject matter. In no case does such identification imply recommendation or endorsement by Securiport LLC, nor does it imply that the equipment identified is necessarily the best available for this purpose. The opinions, findings, and conclusions or recommendations expressed in this publication are those of the authors and do not necessarily reflect the views of our sponsors. 

\bibliographystyle{apalike}
{\small
\bibliography{egbib}}

\section*{\textit{\uppercase{Appendix}}}
\section{Model Training Details}
For the Fourier-based loss functions, the nfft value was set to 5400, giving a frequency resolution of 1 bpm on a 90 frames-per-second (fps) input video. Calibrated models trained with negative samples were trained for 60 epochs. We selected the best-performing model on the DDPM and negative-DDPM validation sets as the final model for testing.

\section{Model Inference Details}
The RPNet model was trained on 90 fps videos from the DDPM dataset, but many of the datasets for testing have lower frame rates. For these videos, we first landmarked and cropped the face as described in ~\cite{Speth_CVIU_2021}, then linearly interpolated the cropped video arrays up to 90 fps. For the CHROM and POS approaches, we estimated the pulse waveforms on the video's native frame rate, then upsampled the waveform with cubic interpolation to 90 samples per second.

\section{Approaches that Failed}

\subsection{Recurrent Neural Networks}
We built several recurrent neural network (RNN) models for anomaly detection. We trained three RNN models by using the feature maps from the seventh, eighth, and ninth layers of pretrained RPNet models as input. We employed a binary cross-entropy loss and trained with both positive and the designed negative samples. The RNN models had one hidden layer and used a gated recurrent unit (GRU). We found that the RNN models did not generalize outside the training data. They gave very accurate predictions when classifying DDPM and the negative DDPM samples, but failed on other datasets. We also found that the RPNet model for input feature maps did not significantly change the performance.

\subsection{Mean Square Error Loss}
The simplest loss we explored was the mean square error (MSE) loss. In this case, we applied the same loss for both negative and positive samples. For positive samples, the target signal was the ground truth waveform. For negative samples, we considered the target sample to be a ``flatline" waveform of zeros. We found that features extracted from this model's predictions were not conducive for anomaly-detection. Furthermore, the MSE loss on positive samples produces worse pulse estimation performance than negative Pearson loss~\cite{Yu2019}.

\end{document}